\documentclass[11pt,a4paper]{article}
\usepackage[hyperref]{naaclhlt2018}
\usepackage{times}
\usepackage{latexsym}
\usepackage{url}

\usepackage{booktabs}
\usepackage{longtable}
\usepackage{wrapfig}
\usepackage{rotating}
\usepackage[normalem]{ulem}
\usepackage{amsmath}
\usepackage{amssymb}
\usepackage{capt-of}
\usepackage{newtxmath}
\usepackage{cleveref}
\usepackage{newtxtext}
\usepackage{mathtools}

\renewcommand{\vec}[1]{\mathbf{#1}}
\newcommand{\vx}{\vec{x}}

\newcommand{\vh}{\vec{h}}
\newcommand{\forward}[1]{#1^f}
\newcommand{\backward}[1]{#1^b}
\newcommand{\vc}{\vec{c}}

\newcommand{\vp}{\vec{p}}
\newcommand{\vv}{\vec{v}}

\newcommand{\R}{\mathbb{R}}
\newcommand{\lossfn}{\mathcal{L}}
\newcommand{\Set}[1]{\mathcal{#1}}
\newcommand{\Sim}{\operatorname{sim}}

\aclfinalcopy 

\title{Learning Visually Grounded Sentence Representations}

\author{
  Douwe Kiela\\
  Facebook AI Research\\
  {\tt dkiela@fb.com} \\\And
  Alexis Conneau \\
  Facebook AI Research\\
  {\tt aconneau@fb.com} \\\AND
  Allan Jabri$\footnotesize ^1$\\
  UC Berkeley\\
  {\tt ajabri@berkeley.edu} \\\And
  Maximilian Nickel\\
  Facebook AI Research\\
  {\tt maxn@fb.com} \\
  }

\date{}

\begin{document}
\maketitle
\addtocounter{footnote}{1}
\footnotetext{Work done while at Facebook AI Research.}

\begin{abstract}
We investigate grounded sentence representations, where we train a sentence encoder to predict the image features of a given caption---i.e., we try to ``imagine'' how a sentence would be depicted visually---and use the resultant features as sentence representations. We examine the quality of the learned representations on a variety of standard sentence representation quality benchmarks, showing improved performance for grounded models over non-grounded ones. In addition, we thoroughly analyze the extent to which grounding contributes to improved performance, and show that the system also learns improved word embeddings.
\end{abstract}

\section{Introduction}

Following the word embedding upheaval of the past few years, one of NLP's next big challenges has become the hunt for universal sentence representations: generic representations of sentence meaning that can be ``plugged into'' any kind of system or pipeline. Examples include Paragraph2Vec \cite{Le:2014icml}, C-Phrase \cite{Pham:2015acl}, SkipThought \cite{Kiros:2015nips} and FastSent \cite{Hill:2016naacl}. These representations tend to be learned from large corpora in an unsupervised setting, much like word embeddings, and effectively ``transferred'' to the task at hand.

Purely text-based semantic models, which represent word
meaning as a distribution over other words
\cite{Harris:1954word,Turney:2010jair,Clark:2015book}, suffer from the grounding problem \cite{Harnad:1990}. It has been shown that grounding leads to improved performance on a variety of word-level tasks \cite{Baroni:2016compass,Kiela:2017thesis}. Unsupervised sentence representation models are often doubly exposed to the grounding problem, especially if they represent sentence meanings as a distribution over other sentences, as in SkipThought \cite{Kiros:2015nips}.

Here, we examine whether grounding also leads to improved sentence representations. In short, the grounding problem is characterized by the lack of an association between symbols and external information. We address this problem by aligning text with paired visual data and hypothesize that sentence representations can be enriched with external information---i.e., grounded---by forcing them to capture visual semantics. We investigate the performance of these representations and the effect of grounding on a variety of semantic benchmarks.

There has been much recent interest in generating actual images from text \cite{Goodfellow:2014nips,Oord:2016nips,Mansimov:2016iclr}. Our method takes a slightly different approach: instead of predicting actual images, we train a deep recurrent neural network to predict the \emph{latent feature representation} of images. That is, we are specifically interested in the semantic content of visual representations and how useful that information is for learning sentence representations. One can think of this as trying to imagine, or form a ``mental picture'', of a sentence's meaning \cite{Chrupala:2015acl}. Much like a sentence's meaning in classical semantics is given by its model-theoretic ground truth \cite{Tarski:1944ppr}, our ground truth is provided by images.

Grounding is likely to be more useful for concrete words and sentences: a sentence such as ``democracy is a political system'' does not yield any coherent mental picture. In order to accommodate the fact that much of language is abstract, we take sentence representations obtained using text-only data (which are better for representing abstract meaning) and combine them with the grounded representations that our system learns (which are good for representing concrete meaning), leading to multi-modal sentence representations.

In what follows, we introduce a system for grounding sentence representations by learning to predict visual content. Although it is not the primary aim of this work, it is important to first examine how well this system achieves what it is trained to do, by evaluating on the COCO5K image and caption retrieval task. We then analyze the performance of grounded representations on a variety of sentence-level semantic transfer tasks, showing that grounding increases performance over text-only representations. We then investigate an important open question in multi-modal semantics: to what extent are improvements in semantic performance due to grounding, rather than to having more data or data from a different distribution? In the remainder, we analyze the role that concreteness plays in representation quality and show that our system learns grounded word embedding projections that outperform non-grounded ones. To the best of our knowledge, this is the first work to comprehensively study grounding for distributed sentence representations on such a wide set of semantic benchmark tasks.

\section{Related work}

\paragraph{Sentence representations} Although there appears to be a consensus
with regard to the methodology for learning word representations, this is much
more of an open problem for sentence representations. Recent work has ranged
from trying to learn to compose word embeddings
\cite{Le:2014icml,Pham:2015acl,Wieting:2016iclr,Arora:2017iclr}, to neural
architectures for predicting the previous and next sentences
\cite{Kiros:2015nips} or learning representations via large-scale supervised tasks \cite{Conneau:2017emnlp}. In particular, SkipThought \cite{Kiros:2015nips} led to an increased interest in learning sentence representations. \newcite{Hill:2016naacl} compare a wide selection of
unsupervised and supervised methods, including a basic caption
prediction system that is similar to ours. That study finds that ``different learning methods are
preferable for different intended applications'', i.e., that the matter of
optimal universal sentence representations is as of yet far from decided.

InferSent \cite{Conneau:2017emnlp} recently showed that supervised sentence representations can be of very high quality. Here, we learn grounded sentence representations in a supervised setting, combine them with standard unsupervised sentence representations, and show how grounding can help for a variety of sentence-level tasks.

\paragraph{Multi-modal semantics} Language grounding in semantics has been motivated by evidence that human meaning representations are grounded in perceptual experience \cite{Jones:1991cd,Perfetti:1998dp,Andrews:2009pr,Riordan:2011tcs}. That is, despite ample evidence of humans representing meaning with respect to an external environment and sensorimotor experience \cite{Barsalou:2008arp,Louwerse:2011tcs}, standard semantic models rely solely on textual data. This gives rise to an infinite regress in text-only semantic representations, i.e., words are defined in terms of other words, \emph{ad infinitum}.

The field of multi-modal semantics, which aims to address this issue by enriching textual representations with information from other modalities, has mostly been concerned with word representations \cite[][and references therein]{Bruni:2014jair,Baroni:2016compass,Kiela:2017thesis}. Learning multi-modal representations that ground text-only representations has been shown to improve performance on a variety of core NLP tasks. This work is most closely related to that of \newcite{Chrupala:2015acl}, who also aim to ground language by relating images to captions: here, we additionally address abstract sentence meaning; have a different architecture, loss function and fusion strategy; and explicitly focus on grounded universal sentence representations.

\paragraph{Bridging vision and language} There is a large body of work that involves jointly embedding images and text, at the word level \cite{Frome:2013nips,Joulin:2016eccv}, the phrase level \cite{Karpathy:2014nips,Li:2016arxiv}, and the sentence level \cite{Karpathy:2015cvpr,Klein:2015cvpr,Kiros:2015nips,Chen:2015cvpr,Reed:2016cvpr}. Our model similarly learns to map sentence representations to be consistent with a visual semantic space, and we focus on studying how these grounded text representations transfer to NLP tasks.

Moreover, there has been a lot of work in recent years on the task of image caption generation \cite{Bernardi:2016jair,Vinyals:2015cvpr,Mao:2015iclr,Fang:cvpr2015}. Here, we do the opposite: we predict the correct image (features) from the caption, rather than the caption from the image (features). Similar ideas were recently successfully applied to multi-modal machine translation \cite{Elliott:2017arxiv,Gella:2017emnlp,Lee:2017arxiv}. Recently, \newcite{Das:2017cvpr} trained dialogue agents to communicate about images, trying to predict image features as well.

\begin{figure*}[t]
    \centering
	\includegraphics[width=0.9\textwidth]{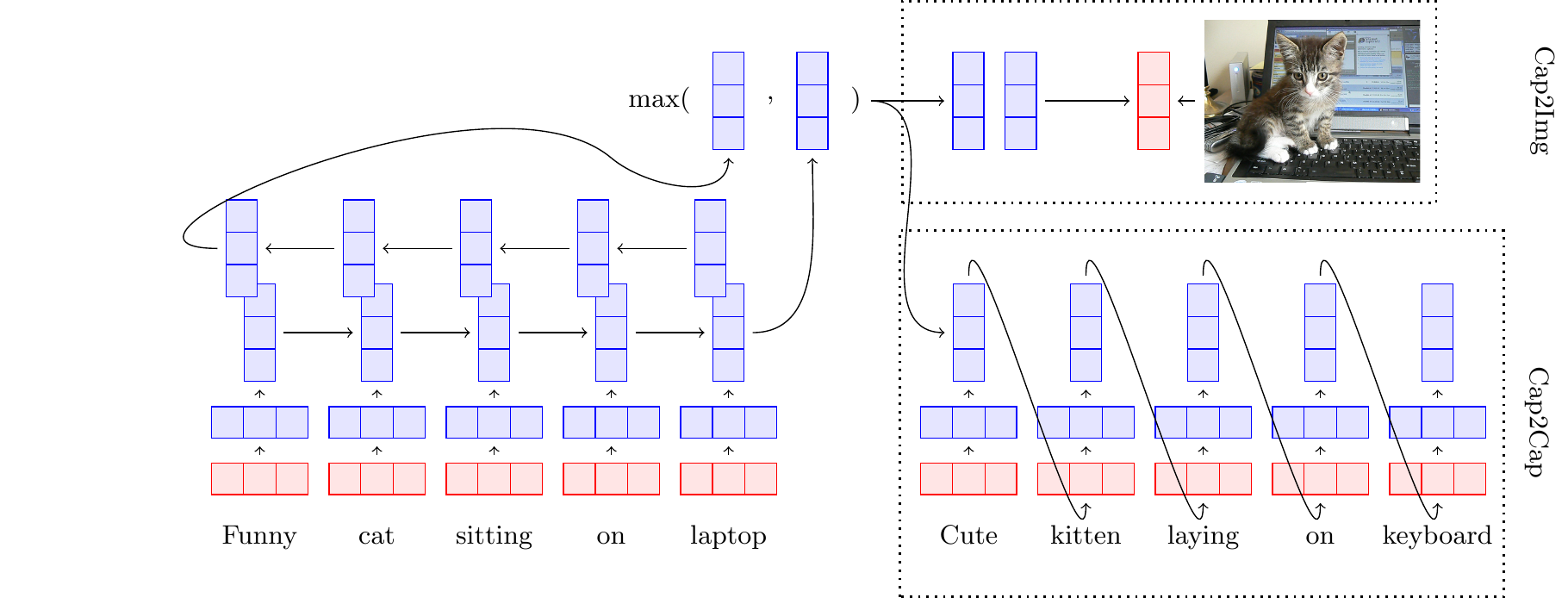}
	\caption{\label{fig:model}Model architecture: predicting either an image (Cap2Img), an alternative caption (Cap2Cap), or both at the same time (Cap2Both).}
\end{figure*}

\section{Approach}
In the following, let \(\Set{D} = \left\{(I_k, \Set{C}_k)\right\}_{k=1}^N\) be a dataset where each image
\(I_k\) is associated with one or more captions \(\Set{C}_k = \{C_1,\ldots,C_{|\Set{C}|_k}\}\).
A prominent example of such a dataset is COCO \cite{Lin:2014eccv}, which
consists of images with up to 5 corresponding captions for each image.
The objective of our approach is to encode a given sentence, i.e., a caption $C$,
and learn to ground it in the corresponding image $I$. To encode the sentence, we train a bidirectional LSTM
(BiLSTM) on the caption, where the input is a sequence of projected word embeddings. We combine the final left-to-right and right-to-left hidden states of the LSTM and take the element-wise maximum to obtain a sentence encoding. We then examine three distinct methods for grounding the sentence encoding.

In the first method, we try to predict the image features (Cap2Img). That is, we learn to map the caption to the same space as the image features that represent the correct image. We call this strong perceptual grounding, where we take the visual input directly into account.

An alternative method is to exploit the fact that one image in COCO has multiple captions (Cap2Cap), and to learn to predict which other captions are valid descriptions of the same image. This approach is strictly speaking not perceptually grounded, but exploits the fact that there is an implicit association between the captions and the shared underlying image, and so could be considered a weaker version of grounding.

Finally, we experiment with a model that optimizes both these objectives jointly: that is, we predict both images and alternative captions for the same image (Cap2Both). Thus, Cap2Both incorporates both strong perceptual and weak implicit grounding. Please see Figure \ref{fig:model} for an illustration of the various models. In what follows, we discuss them in more technical detail.

\subsection{Bidirectional LSTM}

To learn sentence representations, we employ a bidirectional LSTM architecture. In particular, let \({x = (x_1, \ldots, x_T)}\) be an input sequence where
each word is represented via an embedding ${\vx_t \in \R^n}$. Using a standard LSTM \cite{Hochreiter:1997nc}, the hidden state at time \(t\), denoted \(\vh_t \in \R^m\), is computed via

\begin{equation*}
\vh_{t+1}, \vc_{t+1} = \text{LSTM}(\vx_t, \vh_{t}, \vc_{t}\ | \ \Theta)
\end{equation*}

where \(\vc_t\) denotes the cell state of the LSTM and where \(\Theta\) denotes its parameters.

To exploit contextual information in both input directions, we process input
  sentences using a bidirectional LSTM, that reads an input sequence in both normal and reverse order. In particular, for an input sequence \(x\) of
  length \(T\), we compute the hidden state at time \(t\), \(\vh_t \in \R^{2m}\) via
\begin{align*}
  \forward{\vh}_{t+1} & = \text{LSTM}(\vx_t, \forward{\vh}_{t}, \forward{\vc}_{t}\ |\ \forward{\Theta}) \\
  \backward{\vh}_{t+1} & = \text{LSTM}(\vx_{T- t}, \backward{\vh}_{t}, \backward{\vc}_{t}\ |\ \backward{\Theta})
\end{align*}

\noindent Here, the two LSTMs process \(x\) in a forward and a backward order, respectively. We subsequently use \({\max: \R^d \times \R^d \to \R^{d}}\) to combine them into their element-wise maximum, yielding the representation of a caption after it has been processed with the BiLSTM:

\begin{align*}
  \vh_T & = \max(\forward{\vh}_t, \backward{\vh}_{t})
\end{align*}

We use GloVe vectors \cite{Pennington:2014emnlp} for our word embeddings. The embeddings are kept fixed during training, which allows a trained sentence encoder to transfer to tasks (and a vocabulary) that it has not yet seen, provided GloVe embeddings are available. Since GloVe representations are not tuned to represent grounded information, we learn a global transformation of GloVe space to grounded word space. Specifically, let $\overline{\vx} \in \R^n$ be the original GloVe embeddings.
We then learn a linear map $U \in \R^{n \times n}$ such that $\vx = U \overline{\vx}$ and use $\vx$ as input to the BiLSTM. The linear map $U$ and the BiLSTM are trained jointly.

\subsection{Cap2Img}
\label{sec:org074dfe0}
Let \(\vv \in \R^I\) be the latent representation of an image (e.g.the final layer of a ResNet). To ground captions in the images that they describe, we map \(\vh_T\) into the 
latent space of image representations such that their similarity is maximized.
In other words, we aim to predict the latent features of an image from its caption. 
The mapping of caption to image space is performed via a series of projections
\begin{align*}
\vp_{0\phantom{+1}} & = \vh_T\\
\vp_{\ell+1} & = \psi(P_\ell\vp_\ell)
\end{align*}
where \(\psi\) denotes a non-linearity such as ReLUs or tanh.

By jointly training the BiLSTM with these latent projections, we can
then ground the language model in its visual counterpart.
In particular, let $\Theta = \Theta_{\text{BiLSTM}} \cup \{P_\ell\}_{\ell=1}^L$ be the
parameters of the BiLSTM as well as the projection layers.
We then minimize the following ranking loss:
\begin{equation}
  \label{eq:loss:ci}
  \lossfn_{\text{C2I}}(\Theta) = \smashoperator{\sum_{(I, C) \in \Set{D}}} f_\text{rank}(I, C) + f_\text{rank}(C, I)
\end{equation}
where
\begin{equation*}
  f_\text{rank}(a,b) = \sum_{b^\prime \in \Set{N}_a}\left[\gamma - \Sim(a,b) + \Sim(a, b^\prime)\right]_+  
\end{equation*}

where \([x]_+ = \max(0, x)\) denotes the threshold
function at zero and $\gamma$ defines the margin. Furthermore, $\Set{N}_a$ denotes the set of negative samples for an image or
caption and \(\Sim(\cdot,\cdot)\) denotes a similarity measure between vectors. 
In the following, we employ the cosine similarity, i.e.,
\begin{equation*}
\Sim(a, b) = \frac{\langle\vec{a}, \vec{b} \rangle}{\|\vec{a}\| \|\vec{b}\|}.
\end{equation*}

\noindent Although this loss is not smooth at zero, it can be trained end-to-end using subgradient methods. Compared to e.g. an $l_2$ regression loss, \Cref{eq:loss:ci} is less susceptible to error incurred by subspaces of the visual representation that are irrelevant to the high level visual semantics. Empirically, we found it to be more robust to overfitting.

\subsection{Cap2Cap}
\label{sec:orgaacefc7}
Let \(x = (x_1, \ldots, x_T)\), \(y = (y_1, \ldots, y_S)\) be a caption pair
that describes the same image. To learn weakly grounded
  representations, we
employ a standard sequence-to-sequence model \cite{Sutskever:2014nips}, whose task is to predict \(y\)
from \(x\). As in the Cap2Cap model, let $\vh_T$ be the representation
of the input sentence after it has been processed with a BiLSTM. We
then model the joint probability of \(y\) given \(x\) as
\begin{equation*}
p\left(y\ |\ x\right) = \prod_{s=1}^{S} p\left(y_s\ |\ \vh_T, y_1, \ldots, y_{s-1}, \Theta \right).
\end{equation*}
To model the conditional probability of $y_s$ we use the usual multiclass classification approach
over the vocabulary of the corpus $\Set{V}$ such that
\begin{equation*}
  p(y_s = k\ |\ \vh_T, y_1, \ldots, y_{s-1}, \Theta) = \frac{e^{\langle \vv_k, \vec{y}_s \rangle}}{\sum_{j=1}^{|\Set{V}|} e^{\langle \vv_j, \vec{y}_s \rangle}}.
\end{equation*}
Here, $\vec{y}_s = \psi(W_V\vec{g}_s + \vec{b})$ and $\vec{g}_s$ is hidden
state of the decoder LSTM at time $s$.

\noindent To learn the model parameters, we minimize the negative log-likelihood
over all caption pairs, i.e.,

\begin{equation*}
\lossfn_{\text{C2C}}(\theta) = - \smashoperator[l]{\sum_{x,y \in \Set{D}}}\sum_{s=1}^{|y|} \log p(y_s |\ \vh_T, y_1,\ldots,y_{s-1}, \Theta) .
\end{equation*}

\subsection{Cap2Both}
\label{sec:org23ec171}
Finally, we also integrate both concepts of grounding into a joint model, where we optimize the following loss function:

\begin{equation*}
\lossfn_{\text{C2B}}(\Theta) = \lossfn_{C2I}(\Theta) + \lossfn_{C2C}(\Theta) .
\end{equation*}

\subsection{\label{sec:grounded-reps}Grounded universal representations}

On their own, features from this system are likely to suffer from the fact that training on COCO introduces biases: aside from the inherent dataset bias in COCO itself, the system will only have coverage for concrete concepts. COCO is also a much smaller dataset than e.g. the Toronto Books Corpus often used in purely text-based methods \cite{Kiros:2015nips}. As such, grounded representations are potentially less ``universal'' than text-based alternatives, which also cover abstract concepts.

There is evidence that meaning is dually coded in the human brain: while abstract concepts are processed in linguistic areas, concrete concepts are processed in both linguistic and visual areas \cite{Paivio:1990book}. \newcite{Anderson:2017tacl} recently corroborated this hypothesis using semantic representations and fMRI studies. In our case, we want to be able to accommodate concrete sentence meanings, for which our vision-centric system is likely to help; as well as abstract sentence meanings, where trying to ``imagine'' what ``democracy is a political system'' might look like will probably only introduce noise.

Hence, we optionally complement our systems' representations with more abstract universal sentence representations trained on language-only data (specifically, the Toronto Books Corpus). Although it would be interesting to examine multitask scenarios where these representations are jointly learned, we leave this for future work. Here, instead, we combine grounded and language-only representations using simple concatenation, i.e., $r_{gs} = r_{grounded} \mid\mid r_{ling-only}$. Concatenation has been proven to be a strong and straightforward mid-level multi-modal fusion method, previously explored in multi-modal semantics for word representations \cite{Bruni:2014jair,Kiela:2014emnlp}. We call the combined system GroundSent (GS), and distinguish between sentences perceptually grounded in images (GroundSent-Img), weakly grounded in captions (GroundSent-Cap) or grounded in both (GroundSent-Both).

\subsection{Implementation details}

We use 300-dimensional GloVe \cite{Pennington:2014emnlp} embeddings, trained on WebCrawl, for the initial word representations and optimize using Adam \cite{Kingma:2015iclr}. We use ELU \cite{Clevert:2016iclr} for the non-linearity in projection layers, set dropout to 0.5 and use a dimensionality of 1024 for the LSTM. The network was initialized with orthogonal matrices for the recurrent layers \cite{Saxe:2014iclr} and He initialization \cite{He:2015cvpr} for all other layers. The learning rate and margin were tuned on the validation set using grid search.

\begin{table*}[t]
	\centering
	\begin{tabular}{l|lllll|lllll}
		\toprule
		& \multicolumn{10}{c}{COCO5K} \\
		& \multicolumn{5}{c}{Caption Retrieval} & \multicolumn{5}{c}{Image Retrieval} \\
		\textbf{Model} & \textbf{R@1} & \textbf{R@5} & \textbf{R@10} & \textbf{MEDR} & \textbf{MR} & \textbf{R@1} & \textbf{R@5} & \textbf{R@10} & \textbf{MEDR} & \textbf{MR}\\\midrule
		DVSA & 11.8 & 32.5 & 45.4 & 12.2 & NA & 8.9 & 24.9 & 36.3 & 19.5 & NA\\
		FV & 17.3 & 39.0 & 50.2 & 10.0 & 46.4 & 10.8 & 28.3 & 40.1 & 17.0 & 49.3\\
		OE & 23.3 & NA & 65.0 & 5.0 & 24.4 & 18.0 & NA & 57.6 & 7.0 & 35.9\\\midrule
		Cap2Both & 19.4 & 45.0 & 59.4 & 7.0 & 26.5 & 11.7 & 32.6 & 46.4 & 12.0 & 41.7\\
		Cap2Img & 27.1 & 55.6 & 70.0 & 4.0 & 19.2 & 17.1 & 43.0 & 57.3 & 8.0 & 36.6\\
		\bottomrule
	\end{tabular}
	\caption{\label{table:coco}Retrieval (higher is better) results on COCO, plus median rank (MEDR) and mean rank (MR) (lower is better). Note that while this work underwent review, better methods have been published, most notably VSE++ \cite{Faghri:2017arxiv}.}
\end{table*}

\begin{table*}[t]
	\centering
	\begin{tabular}{l|llllll|ll}
		\toprule
		\textbf{Model} & \textbf{MR} & \textbf{CR} & \textbf{SUBJ} & \textbf{MPQA} & \textbf{MRPC} & \textbf{SST} & \textbf{SNLI} & \textbf{SICK}\\\midrule
		ST-LN & 78.1 & 80.1 & 92.7 & 88.0 & 69.6/81.2 & 82.9 & 73.8 & 78.5\\\midrule
		GroundSent-Cap & 79.9 & 81.4 & 93.1 & 88.9 & 72.9/82.2 & 85.0 & 75.5 & 79.7\\
		GroundSent-Img & 79.1 & 80.8 & 93.1 & 89.0 & 71.9/81.4 & 86.1 & 76.1 & 82.2\\
		GroundSent-Both & 79.6 & 81.7 & 93.4 & 89.4 & 72.7/82.5 & 84.8 & 76.1 & 81.6\\
		\bottomrule
	\end{tabular}
	\caption{\label{table:semantic-transfer}Accuracy results on sentence classification and entailment tasks. 
	}
\end{table*}

\section{Data, evaluation and comparison}

We use the same COCO splits as \newcite{Karpathy:2015cvpr} for training (113,287 images), validation (5000 images) and testing (5000 images).  Image features for COCO were obtained by transferring the final layer from a ResNet-101 \cite{He:2016cvpr} trained on ImageNet (ILSVRC 2015).

\subsection{Transfer tasks}

We are specifically interested in how well (grounded) universal sentence representations transfer to different tasks. To evaluate this, we perform experiments for a variety of tasks. In all cases, we compare against layer-normalized Skip-Thought vectors, a well-known high-performing sentence encoding method \cite{Ba:2016arxiv}. To ensure that we use the exact same evaluations, with identical hyperparameters and settings, we evaluate all systems with the same evaluation pipeline, namely SentEval \cite{Conneau:2018lrec}\footnote{See https://github.com/facebookresearch/SentEval. The aim of SentEval is to encompass a  comprehensive set of benchmarks that has been loosely established in the research community as the standard for evaluating sentence representations.}. Following previous work in the field, the idea is to take universal sentence representations and to learn a simple classifier on top for each of the transfer tasks---the higher the quality of the sentence representation, the better the performance on these transfer tasks should be.

\subsubsection{Semantic classification}

We evaluate on the following well-known and widely used evaluations: movie review sentiment (MR) \cite{Pang:2005acl}, product reviews (CR) \cite{Hu:2004kdd}, subjectivity classification (SUBJ) \cite{Pang:2004acl}, opinion polarity (MPQA) \cite{Wiebe:2005lrec}, paraphrase identification (MSRP) \cite{Dolan:2004acl} and sentiment classification (SST, binary version) \cite{Socher:2013emnlp}. Accuracy is measured in all cases, except for MRPC, which measures accuracy and the F1-score.

\subsubsection{Entailment}

Recent years have seen an increased interest in entailment classification as an appropriate evaluation of sentence representation quality. We evaluate representations on two well-known entailment, or natural language inference, datasets: the large-scale SNLI dataset \cite{Bowman:2015emnlp} and the SICK dataset \cite{Marelli:2014lrec}.

\subsection{Implementational details}

We implement a simple logistic regression on top of the sentence representation. In the cases of SNLI and SICK, as is the standard for these datasets, the representations for the individual sentences $u$ and $v$ are combined by using $\langle \vec{u}, \vec{v}, \vec{u}*\vec{v}, |\vec{u}-\vec{v}| \rangle$ as the input features. We tune the seed and an $l_2$ penalty on the validation sets for each, and train using Adam \cite{Kingma:2015iclr}, with a learning rate of 0.001 and a batch size of 32.

\begin{table*}
	\centering
	\begin{tabular}{l|llllll|ll}
		\toprule
		\textbf{Model} & \textbf{MR} & \textbf{CR} & \textbf{SUBJ} & \textbf{MPQA} & \textbf{MRPC} & \textbf{SST} & \textbf{SNLI} & \textbf{SICK}\\\midrule

STb-1024	& 70.3 & 68.0 & 87.5 & 85.5 & 69.7/80.6 & 78.3 & 67.3 & 76.6\\\midrule

STb-2048		& 73.1 & \textbf{75.7} & 88.3 & 86.5 & 71.6/\textbf{81.7} & 79.0 & 71.0 & 78.8\\
2$\times$STb-1024		& 71.4 & 74.7 & 88.2 & 86.6 & 71.3/80.7 & 75.8 & 69.4 & 78.3\\\midrule

Cap2Cap		& 71.4 & 74.7 & 86.7 & \textbf{86.7} & 70.3/79.8 & 76.1 & 68.5 & 78.2\\
Cap2Img		& 72.1 & 75.5 & 86.9 & 86.0 & \textbf{72.3}/81.1 & 77.7 & 71.4 & 81.2\\
Cap2Both	& 71.6 & 74.4 & 86.5 & 85.5 & 71.4/79.5 & 78.5 & 71.3 & \textbf{81.7}\\\midrule

GroundSent-Cap	& 73.1 & 73.0 & \textbf{88.6} & 86.6 & 70.8/81.2 & 79.4 & 70.7 & 79.1\\
GroundSent-Img	& 72.5 & 74.9 & 88.4 & 85.7 & 71.3/81.2 & 79.4 & 70.5 & 79.7\\
GroundSent-Both & \textbf{73.3} & 75.2 & 87.5 & 86.6 & 69.9/79.9 & \textbf{80.3} & \textbf{72.0} & 78.1\\

		\bottomrule
	\end{tabular}
	\caption{\label{table:grounding-contribution}Thorough investigation of the contribution of grounding, ensuring equal number of components and identical architectures, on the variety of sentence-level semantic benchmark tasks. STb=SkipThought-like model with bidirectional LSTM+max. 2$\times$STb-1024=ensemble of 2 different STb models with different initializations. GroundSent is STb-1024+Cap2Cap/Img/Both. We find that performance improvements are sometimes due to having more parameters, but in most cases due to grounding.}
\end{table*}

\section{Results}

Although it is not the primary aim of this work to learn a state-of-the-art image and caption retrieval system, it is important to first establish the capability of our system to do what it is trained to do. Table \ref{table:coco} shows the results on the COCO5K caption and image retrieval tasks for the two models that predict image features.  We compare our system against several well-known approaches, namely Deep Visual-Semantic Alignments (DVSA) \cite{Karpathy:2015cvpr}, Fisher Vectors (FV) \cite{Klein:2015cvpr} and Order Embeddings (OE) \cite{Vendrov:2015iclr}.
As the results show, Cap2Img performs very well
on this task, outperforming the compared models on caption retrieval and being very close to order embeddings on image retrieval\footnote{In fact, we found that we can achieve better performance on this task by reducing the dimensionality of the encoder. A lower dimensionality in the encoder also reduces the transferability of the features, unfortunately, so we leave a more thorough investigation of this phenomenon for future work.}. The fact that the system outperforms Order Embeddings on caption retrieval suggests that it has a better sentence encoder. Cap2Both does not work as well on this task as the image-only case, probably because interference from the language signal makes the problem harder to optimize. The results indicate that the system has learned to predict image features from captions, and captions from images, at a level exceeding or close to the state-of-the-art on this task.

\subsection{Transfer task performance}

Having established that we can learn high-quality grounded sentence encodings, the core question we now wish to examine is how well grounded sentence representations transfer. In this section, we combine our grounded features with the high-quality layer-normalized SkipThought representations of \newcite{Ba:2016arxiv}, leading to multi-modal sentence representations as described in Section \ref{sec:grounded-reps}. That is, we concatenate Cap2Cap, Cap2Img or Cap2Both and Skip-Thought with Layer Normalization (ST-LN) representations, yielding GroundSent-Cap, GroundSent-Img and GroundSent-Both representations, respectively. We report performance of ST-LN using SentEval, which led to slightly different numbers than what is reported in their paper\footnote{This is probably due to different seeds, optimization methods and other minor implementational details that differ between the original work and SentEval.}.

Table \ref{table:semantic-transfer} shows the results for the semantic classification and entailment tasks. Note that all systems use the exact same evaluation pipeline, which makes them directly comparable. We can see that in all cases, grounding increases the performance. The question of which type of grounding works best is more difficult: generally, grounding with Cap2Cap and Cap2Both appears to do slightly better on most tasks, but on e.g. SST, Cap2Img works better. The entailment task results (SNLI and SICK in Table \ref{table:semantic-transfer}) show a similar picture: in all cases grounding improves performance.

It is important to note that, in this work, we are not necessarily concerned with replacing the state-of-the-art on these tasks: there are systems that perform better. We are primarily interested in whether grounding helps relative to text-only baselines. We find that it does.

\subsection{The contribution of grounding}

An important open question is whether the increase in performance in multi-modal semantic models is due to qualitatively different information from \emph{grounding}, or simply due to the fact that we have \emph{more parameters} or \emph{data from a different distribution}. In order to examine this, we implement a SkipThought-like model that also uses a bidirectional LSTM with element-wise max on the final hidden layer (henceforth referred to as STb). This model is architecturally identical to the sentence encoder used before: it can be thought of as Cap2Cap, but where the objective is not to predict an alternative caption, but to predict the previous and next sentence in the Toronto Books Corpus, just like SkipThought \cite{Kiros:2015nips}.

We train a 1024-dimensional and 2048-dimensional STb model (for one full iteration, with all other hyperparameters identical to Cap2Cap) to compare against: if grounding improves results because it introduces qualitatively different information, rather than just from having more parameters (i.e., a higher embedding dimensionality), we should expect the multi-modal GroundSent models to perform better not only than STb-1024, but also than STb-2048, which has the same number of parameters (recall that GroundSent models are combinations of grounded and linguistic-only representations). In addition, we compare against an ``ensemble'' of two different STb-1024 models (i.e., a concatenation of two separately trained STb-1024), to check that we are not (just) observing an ensemble effect.

As Table \ref{table:grounding-contribution} shows, a more nuanced picture emerges in this comparison: grounding helps more for some datasets than for others. Grounded models outperform the STb-1024 model (which uses much more data---the Toronto Books Corpus is much larger than COCO) in all cases, often already without concatenating the textual modality. The ensemble of two STb-1024 models performs better than the individual one, and so does the higher-dimensional one. In the cases of CR and MRPC (F1), it appears that improved performance is due to having more data or ensemble effects. For the other datasets, grounding clearly yields better results. These results indicate that grounding does indeed capture qualitatively different information, yielding better universal sentence representations.

\section{Discussion}

There are a few other important questions to investigate. The average abstractness or concreteness of the evaluation datasets may have a large impact on performance. In addition, word embeddings from the learned projection from GloVe input embeddings, which now provides a generic word-embedding grounding method even for words that are not present in the image-caption training data, can be examined.

\begin{table}
	\centering
	\begin{tabular}{l|l}
		\toprule
		\textbf{Dataset} & \textbf{Concreteness}\\\midrule
		MR & 2.3737 $\pm$ 0.965\\
		CR & 2.4714 $\pm$ 1.025\\
		SUBJ & 2.4510 $\pm$ 1.007\\
		MPQA & 2.3158 $\pm$ 0.834\\
		MRPC & 2.5086 $\pm$ 0.987\\
		SST	& 2.7471 $\pm$ 1.142\\\midrule
		SNLI & 3.1867 $\pm$ 1.309\\
		SICK & 3.1282 $\pm$ 1.372\\
		\bottomrule
	\end{tabular}
	\caption{\label{table:concreteness}Mean and variance of dataset concreteness, over all words in the datasets.}
\end{table}

\subsection{Concreteness}

As we have seen, performance across datasets and models can vary substantially. A dataset's concreteness plays an important role in the relative merit of applying grounding: a dataset consisting mostly of abstract words is less likely to benefit from grounding than one that uses mostly concrete words. In order to examine this effect, we calculate the average concreteness of the evaluation datasets used in this study. Table \ref{table:concreteness} shows the average human-annotated concreteness ratings for all words (where available) in each dataset. The ratings were obtained by \newcite{Brysbaert:2014brm} in a large-scale study, yielding scores for 40,000 English words.

We observe that the two entailment datasets are more concrete, which is due to the fact that the premises are derived from caption datasets (Flickr30K in the case of SNLI; Flickr8K and video captions in the case of SICK). This explains why grounding can clearly be seen to help in these cases. For the semantic classification tasks, the more concrete datasets are MRPC and SST. The picture is less clear for the first, but in SST we see that the grounded representations definitely do work better. Concreteness values make it easier to analyze performance, but  are apparently not always direct indicators of improvements with grounding.

\subsection{Grounded word embeddings}

Our models contain a projection layer that maps the GloVe word embeddings that they receive as inputs to a different embedding space. There has been a lot of interest in grounded word representations in recent years, so it is interesting to examine what kind of word representations our models learn. We omit Cap2Cap for reasons of space (it performs similarly to Cap2Both). As shown in Table \ref{table:word-embeddings}, the grounded word projections that our network learns yield higher-quality word embeddings on four standard lexical semantic similarity benchmarks: MEN \cite{Bruni:2014jair}, SimLex-999 \cite{Hill:2016cl}, Rare Words \cite{Luong:2013conll} and WordSim-353 \cite{Finkelstein:2001www}.

\begin{table}
	\centering
	\begin{tabular}{l|rrrr}
		\toprule
		\textbf{Model} & \textbf{MEN} & \textbf{SimLex} & \textbf{RW} & \textbf{W353}\\\midrule
		GloVe & 0.805 & 0.408 & 0.451 & 0.738\\\midrule
		Cap2Both & 0.819 & 0.467 & 0.487 & 0.712\\
		Cap2Img & 0.845 & 0.515 & 0.523 & 0.753\\
		\bottomrule
	\end{tabular}
	\caption{\label{table:word-embeddings}Spearman $\rho_s$ correlation on four standard semantic similarity evaluation benchmarks.}
\end{table}

\section{Conclusion}

We have investigated grounding for universal sentence representations. We achieved good performance on caption and image retrieval tasks on the large-scale COCO dataset. We subsequently showed how the sentence encodings that the system learns can be transferred to various NLP tasks, and that grounded universal sentence representations lead to improved performance. We analyzed the source of improvements from grounding, and showed that the increased performance appears to be due to the introduction of qualitatively different information (i.e., grounding), rather than simply having more parameters or applying ensemble methods. Lastly, we showed that our systems learned high-quality grounded word embeddings that outperform non-grounded ones on standard semantic similarity benchmarks. It could well be that our methods are even more suited for more concrete tasks, such as visual question answering, visual storytelling, or image-grounded dialogue---an avenue worth exploring in future work. In addition, it would be interesting to explore multi-task learning for sentence representations where one of the tasks involves grounding.

\section*{Acknowledgments}
We thank the anonymous reviewers for their helpful comments and suggestions. Part of Fig.~\ref{fig:model}~is~licensed~from~dougwoods/CC-BY-2.0/flickr.com/photos/deerwooduk/682390157.

\bibliography{naaclhlt2018}

\begin{thebibliography}{}
\expandafter\ifx\csname natexlab\endcsname\relax\def\natexlab#1{#1}\fi

\bibitem[{Anderson et~al.(2017)Anderson, Kiela, Clark, and
  Poesio}]{Anderson:2017tacl}
Andrew~J. Anderson, Douwe Kiela, Stephen Clark, and Massimo Poesio. 2017.
\newblock Visually grounded and textual semantic models differentially decode
  brain activity associated with concrete and abstract nouns.
\newblock {\em Transactions of the Association for Computational Linguistics\/}
  5.

\bibitem[{Andrews et~al.(2009)Andrews, Vigliocco, and Vinson}]{Andrews:2009pr}
Mark Andrews, Gabriella Vigliocco, and David Vinson. 2009.
\newblock Integrating experiential and distributional data to learn semantic
  representations.
\newblock {\em Psychological review\/} 116(3):463.

\bibitem[{Arora et~al.(2017)Arora, Liang, and Ma}]{Arora:2017iclr}
Sanjeev Arora, Yingyu Liang, and Tengyu Ma. 2017.
\newblock A simple but tough-to-beat baseline for sentence embeddings.
\newblock In {\em International Conference on Learning Representations
  (ICLR)\/}.

\bibitem[{Ba et~al.(2016)Ba, Kiros, and Hinton}]{Ba:2016arxiv}
Jimmy~Lei Ba, Jamie~Ryan Kiros, and Geoffrey~E Hinton. 2016.
\newblock Layer normalization.
\newblock {\em arXiv preprint arXiv:1607.06450\/} .

\bibitem[{Baroni(2016)}]{Baroni:2016compass}
Marco Baroni. 2016.
\newblock Grounding distributional semantics in the visual world.
\newblock {\em Language and Linguistics Compass\/} 10(1):3--13.

\bibitem[{Barsalou(2008)}]{Barsalou:2008arp}
Lawrence~W. Barsalou. 2008.
\newblock Grounded cognition.
\newblock {\em Annual Review of Psychology\/} 59(1):617--645.

\bibitem[{Bernardi et~al.(2016)Bernardi, Cakici, Elliott, Erdem, Erdem,
  Ikizler-Cinbis, Keller, Muscat, and Plank}]{Bernardi:2016jair}
Raffaella Bernardi, Ruket Cakici, Desmond Elliott, Aykut Erdem, Erkut Erdem,
  Nazli Ikizler-Cinbis, Frank Keller, Adrian Muscat, and Barbara Plank. 2016.
\newblock Automatic description generation from images: A survey of models,
  datasets, and evaluation measures.
\newblock {\em Journal of Artificial Intelligence Research (JAIR)\/} pages
  409--442.

\bibitem[{Bowman et~al.(2015)Bowman, Angeli, Potts, and
  Manning}]{Bowman:2015emnlp}
Samuel~R. Bowman, Gabor Angeli, Christopher Potts, and Christopher~D. Manning.
  2015.
\newblock A large annotated corpus for learning natural language inference.
\newblock In {\em Proceedings of EMNLP\/}.

\bibitem[{Bruni et~al.(2014)Bruni, Tran, and Baroni}]{Bruni:2014jair}
Elia Bruni, Nam{-}Khanh Tran, and Marco Baroni. 2014.
\newblock Multimodal distributional semantics.
\newblock {\em Journal of Artifical Intelligence Research\/} 49:1--47.

\bibitem[{Brysbaert et~al.(2014)Brysbaert, Warriner, and
  Kuperman}]{Brysbaert:2014brm}
Marc Brysbaert, Amy~Beth Warriner, and Victor Kuperman. 2014.
\newblock Concreteness ratings for 40 thousand generally known english word
  lemmas.
\newblock {\em Behavior research methods\/} 46(3):904--911.

\bibitem[{Chen and Zitnick(2015)}]{Chen:2015cvpr}
Xinlei Chen and Lawrence~C Zitnick. 2015.
\newblock Mind's eye: A recurrent visual representation for image caption
  generation.
\newblock In {\em Proceedings of the IEEE conference on computer vision and
  pattern recognition\/}. pages 2422--2431.

\bibitem[{Chrupa{\l}a et~al.(2015)Chrupa{\l}a, K\'{a}d\'{a}r, and
  Alishahi}]{Chrupala:2015acl}
Grzegorz Chrupa{\l}a, \'{A}kos K\'{a}d\'{a}r, and Afra Alishahi. 2015.
\newblock Learning language through pictures.
\newblock In {\em Proceedings of ACL\/}.

\bibitem[{Clark(2015)}]{Clark:2015book}
Stephen Clark. 2015.
\newblock Vector {S}pace {M}odels of {L}exical {M}eaning.
\newblock In Shalom Lappin and Chris Fox, editors, {\em Handbook of
  Contemporary Semantic Theory\/}, Wiley-Blackwell, Oxford, chapter~16.

\bibitem[{Clevert et~al.(2016)Clevert, Unterthiner, and
  Hochreiter}]{Clevert:2016iclr}
Djork-Arn{\'e} Clevert, Thomas Unterthiner, and Sepp Hochreiter. 2016.
\newblock Fast and accurate deep network learning by exponential linear units
  ({ELU}s).
\newblock In {\em International Conference on Learning Representations
  (ICLR)\/}.

\bibitem[{Conneau and Kiela(2018)}]{Conneau:2018lrec}
Alexis Conneau and Douwe Kiela. 2018.
\newblock Senteval: An evaluation toolkit for universal sentence
  representations.
\newblock In {\em Proceedings of LREC\/}.

\bibitem[{Conneau et~al.(2017)Conneau, Kiela, Schwenk, Barrault, and
  Bordes}]{Conneau:2017emnlp}
Alexis Conneau, Douwe Kiela, Holger Schwenk, Lo{\"{\i}}c Barrault, and Antoine
  Bordes. 2017.
\newblock Supervised learning of universal sentence representations from
  natural language inference data.
\newblock In {\em Proceedings of EMNLP\/}. Copenhagen, Denmark.

\bibitem[{Das et~al.(2017)Das, Kottur, Moura, Lee, and Batra}]{Das:2017cvpr}
Abhishek Das, Satwik Kottur, Jos{\'{e}} M.~F. Moura, Stefan Lee, and Dhruv
  Batra. 2017.
\newblock Learning cooperative visual dialog agents with deep reinforcement
  learning.
\newblock In {\em Proceedings of CVPR\/}.

\bibitem[{Dolan et~al.(2004)Dolan, Quirk, and Brockett}]{Dolan:2004acl}
Bill Dolan, Chris Quirk, and Chris Brockett. 2004.
\newblock Unsupervised construction of large paraphrase corpora: Exploiting
  massively parallel news sources.
\newblock In {\em Proceedings of ACL\/}. page 350.

\bibitem[{Elliott and K{\'a}d{\'a}r(2017)}]{Elliott:2017arxiv}
Desmond Elliott and {\'A}kos K{\'a}d{\'a}r. 2017.
\newblock Imagination improves multimodal translation.
\newblock {\em arXiv preprint arXiv:1705.04350\/} .

\bibitem[{Faghri et~al.(2017)Faghri, Fleet, Kiros, and
  Fidler}]{Faghri:2017arxiv}
Fartash Faghri, David~J Fleet, Jamie~Ryan Kiros, and Sanja Fidler. 2017.
\newblock Vse++: Improved visual-semantic embeddings.
\newblock {\em arXiv preprint arXiv:1707.05612\/} .

\bibitem[{Fang et~al.(2015)Fang, Gupta, Iandola, Srivastava, Deng, Dollar, Gao,
  He, Mitchell, Platt, Zitnick, and Zweig}]{Fang:cvpr2015}
H.~Fang, S.~Gupta, F.N. Iandola, R.~Srivastava, L.~Deng, P.~Dollar, J.~Gao,
  X.~He, M.~Mitchell, J.C. Platt, C.L. Zitnick, and G.~Zweig. 2015.
\newblock From captions to visual concepts and back.
\newblock In {\em CVPR\/}.

\bibitem[{Finkelstein et~al.(2001)Finkelstein, Gabrilovich, Matias, Rivlin,
  Solan, Wolfman, and Ruppin}]{Finkelstein:2001www}
Lev Finkelstein, Evgeniy Gabrilovich, Yossi Matias, Ehud Rivlin, Zach Solan,
  Gadi Wolfman, and Eytan Ruppin. 2001.
\newblock Placing search in context: The concept revisited.
\newblock In {\em Proceedings of the 10th international conference on World
  Wide Web (WWW)\/}. ACM, pages 406--414.

\bibitem[{Frome et~al.(2013)Frome, Corrado, Shlens, Bengio, Dean, and
  Mikolov}]{Frome:2013nips}
A.~Frome, G.~Corrado, J.~Shlens, S.~Bengio, J.~Dean, and T.~Mikolov. 2013.
\newblock Devise: A deep visual-semantic embedding model.
\newblock In {\em NIPS\/}.

\bibitem[{Gella et~al.(2017)Gella, Sennrich, Keller, and
  Lapata}]{Gella:2017emnlp}
Spandana Gella, Rico Sennrich, Frank Keller, and Mirella Lapata. 2017.
\newblock Image pivoting for learning multilingual multimodal representations .

\bibitem[{Goodfellow et~al.(2014)Goodfellow, Pouget-Abadie, Mirza, Xu,
  Warde-Farley, Ozair, Courville, and Bengio}]{Goodfellow:2014nips}
Ian Goodfellow, Jean Pouget-Abadie, Mehdi Mirza, Bing Xu, David Warde-Farley,
  Sherjil Ozair, Aaron Courville, and Yoshua Bengio. 2014.
\newblock Generative adversarial nets.
\newblock In {\em Proceedings of NIPS\/}.

\bibitem[{Harnad(1990)}]{Harnad:1990}
Stevan Harnad. 1990.
\newblock The symbol grounding problem.
\newblock {\em Physica D\/} 42:335--346.

\bibitem[{Harris(1954)}]{Harris:1954word}
Z.~Harris. 1954.
\newblock {Distributional Structure}.
\newblock {\em Word\/} 10(23):146---162.

\bibitem[{He et~al.(2015)He, Zhang, Ren, and Sun}]{He:2015cvpr}
Kaiming He, Xiangyu Zhang, Shaoqing Ren, and Jian Sun. 2015.
\newblock Delving deep into rectifiers: Surpassing human-level performance on
  imagenet classification.
\newblock In {\em Proceedings of the IEEE international conference on computer
  vision (CVPR)\/}. pages 1026--1034.

\bibitem[{He et~al.(2016)He, Zhang, Ren, and Sun}]{He:2016cvpr}
Kaiming He, Xiangyu Zhang, Shaoqing Ren, and Jian Sun. 2016.
\newblock Deep residual learning for image recognition.
\newblock In {\em Proceedings of the IEEE Conference on Computer Vision and
  Pattern Recognition (CVPR)\/}. pages 770--778.

\bibitem[{Hill et~al.(2016{\natexlab{a}})Hill, Cho, and
  Korhonen}]{Hill:2016naacl}
Felix Hill, Kyunghyun Cho, and Anna Korhonen. 2016{\natexlab{a}}.
\newblock Learning distributed representations of sentences from unlabelled
  data.
\newblock In {\em Proceedings of NAACL\/}.

\bibitem[{Hill et~al.(2016{\natexlab{b}})Hill, Reichart, and
  Korhonen}]{Hill:2016cl}
Felix Hill, Roi Reichart, and Anna Korhonen. 2016{\natexlab{b}}.
\newblock Simlex-999: Evaluating semantic models with (genuine) similarity
  estimation.
\newblock {\em Computational Linguistics\/} .

\bibitem[{Hochreiter and Schmidhuber(1997)}]{Hochreiter:1997nc}
Sepp Hochreiter and J{\"u}rgen Schmidhuber. 1997.
\newblock Long short-term memory.
\newblock {\em Neural computation\/} 9(8):1735--1780.

\bibitem[{Hu and Liu(2004)}]{Hu:2004kdd}
Minqing Hu and Bing Liu. 2004.
\newblock Mining and summarizing customer reviews.
\newblock In {\em Proceedings of SIGKDD\/}. pages 168--177.

\bibitem[{Jones et~al.(1991)Jones, Smith, and Landau}]{Jones:1991cd}
Susan~S Jones, Linda~B Smith, and Barbara Landau. 1991.
\newblock Object properties and knowledge in early lexical learning.
\newblock {\em Child development\/} 62(3):499--516.

\bibitem[{Joulin et~al.(2016)Joulin, van~der Maaten, Jabri, and
  Vasilache}]{Joulin:2016eccv}
A.~Joulin, L.J.P. van~der Maaten, A.~Jabri, and N.~Vasilache. 2016.
\newblock Learning visual features from large weakly supervised data.
\newblock In {\em ECCV\/}.

\bibitem[{Karpathy et~al.(2014)Karpathy, Joulin, and
  Fei-Fei}]{Karpathy:2014nips}
A.~Karpathy, A.~Joulin, and L.~Fei-Fei. 2014.
\newblock Deep fragment embeddings for bidirectional image sentence mapping.
\newblock In {\em Proceedings of NIPS\/}.

\bibitem[{Karpathy and Fei-Fei(2015)}]{Karpathy:2015cvpr}
Andrej Karpathy and Li~Fei-Fei. 2015.
\newblock Deep visual-semantic alignments for generating image descriptions.
\newblock In {\em Proceedings of the IEEE Conference on Computer Vision and
  Pattern Recognition (CVPR)\/}. pages 3128--3137.

\bibitem[{Kiela(2017)}]{Kiela:2017thesis}
Douwe Kiela. 2017.
\newblock {Deep embodiment: grounding semantics in perceptual modalities (PhD
  thesis)}.
\newblock Technical Report UCAM-CL-TR-899, University of Cambridge, Computer
  Laboratory.

\bibitem[{Kiela and Bottou(2014)}]{Kiela:2014emnlp}
Douwe Kiela and L\'{e}on Bottou. 2014.
\newblock {Learning Image Embeddings using Convolutional Neural Networks for
  Improved Multi-Modal Semantics}.
\newblock In {\em {Proceedings of EMNLP}\/}.

\bibitem[{Kingma and Ba(2015)}]{Kingma:2015iclr}
Diederik Kingma and Jimmy Ba. 2015.
\newblock Adam: A method for stochastic optimization.
\newblock In {\em International Conference on Learning Representations
  (ICLR)\/}.

\bibitem[{Kiros et~al.(2015)Kiros, Zhu, Salakhutdinov, Zemel, Torralba,
  Urtasun, and Fidler}]{Kiros:2015nips}
Ryan Kiros, Yukun Zhu, Ruslan Salakhutdinov, Richard~S. Zemel, Antonio
  Torralba, Raquel Urtasun, and Sanja Fidler. 2015.
\newblock Skip-thought vectors.
\newblock In {\em Proceedings of NIPS\/}.

\bibitem[{Klein et~al.(2015)Klein, Lev, Sadeh, and Wolf}]{Klein:2015cvpr}
Benjamin Klein, Guy Lev, Gil Sadeh, and Lior Wolf. 2015.
\newblock Associating neural word embeddings with deep image representations
  using fisher vectors.
\newblock In {\em Proceedings of the IEEE Conference on Computer Vision and
  Pattern Recognition (CVPR)\/}. pages 4437--4446.

\bibitem[{Le and Mikolov(2014)}]{Le:2014icml}
Quoc~V. Le and Tomas Mikolov. 2014.
\newblock Distributed representations of sentences and documents.
\newblock In {\em Proceedings of ICML\/}.

\bibitem[{Lee et~al.(2017)Lee, Cho, Weston, and Kiela}]{Lee:2017arxiv}
Jason Lee, Kyunghyun Cho, Jason Weston, and Douwe Kiela. 2017.
\newblock Emergent translation in multi-agent communication.
\newblock {\em CoRR\/} abs/1710.06922.

\bibitem[{Li et~al.(2016)Li, Jabri, Joulin, and van~der Maaten}]{Li:2016arxiv}
A.~Li, A.~Jabri, A.~Joulin, and L.J.P. van~der Maaten. 2016.
\newblock Learning visual n-grams from web data.
\newblock In {\em arxiv\/}.

\bibitem[{Lin et~al.(2014)Lin, Maire, Belongie, Hays, Perona, Ramanan,
  Doll{\'a}r, and Zitnick}]{Lin:2014eccv}
Tsung-Yi Lin, Michael Maire, Serge Belongie, James Hays, Pietro Perona, Deva
  Ramanan, Piotr Doll{\'a}r, and C~Lawrence Zitnick. 2014.
\newblock Microsoft {COCO}: {C}ommon objects in context.
\newblock In {\em European Conference on Computer Vision (ECCV)\/}. Springer,
  pages 740--755.

\bibitem[{Louwerse(2008)}]{Louwerse:2011tcs}
Max~M. Louwerse. 2008.
\newblock Symbol interdependency in symbolic and embodied cognition.
\newblock {\em Topics in Cognitive Science\/} 59(1):617--645.

\bibitem[{Luong et~al.(2013)Luong, Socher, and Manning}]{Luong:2013conll}
Thang Luong, Richard Socher, and Christopher~D Manning. 2013.
\newblock Better word representations with recursive neural networks for
  morphology.
\newblock In {\em Proceedings of CoNLL\/}. pages 104--113.

\bibitem[{Mansimov et~al.(2016)Mansimov, Parisotto, Ba, and
  Salakhutdinov}]{Mansimov:2016iclr}
Elman Mansimov, Emilio Parisotto, Jimmy~Lei Ba, and Ruslan Salakhutdinov. 2016.
\newblock Generating images from captions with attention.
\newblock In {\em International Conference on Learning Representations
  (ICLR)\/}.

\bibitem[{Mao et~al.(2015)Mao, Xu, Yang, Wang, and Yuille}]{Mao:2015iclr}
J.~Mao, W.~Xu, Y.~Yang, J.~Wang, and A.L. Yuille. 2015.
\newblock Deep captioning with multimodal recurrent neural networks.
\newblock In {\em Proceedings of ICLR\/}.

\bibitem[{Marelli et~al.(2014)Marelli, Menini, Baroni, Bentivogli, Bernardi,
  and Zamparelli}]{Marelli:2014lrec}
Marco Marelli, Stefano Menini, Marco Baroni, Luisa Bentivogli, Reffaella
  Bernardi, and Roberto Zamparelli. 2014.
\newblock A {SICK} cure for the evaluation of compositional distributional
  semantic models.
\newblock In {\em Proceedings of LREC\/}.

\bibitem[{Paivio(1990)}]{Paivio:1990book}
Allan Paivio. 1990.
\newblock {\em Mental representations: A dual coding approach\/}.
\newblock Oxford University Press.

\bibitem[{Pang and Lee(2004)}]{Pang:2004acl}
Bo~Pang and Lillian Lee. 2004.
\newblock A sentimental education: Sentiment analysis using subjectivity
  summarization based on minimum cuts.
\newblock In {\em Proceedings of ACL\/}. page 271.

\bibitem[{Pang and Lee(2005)}]{Pang:2005acl}
Bo~Pang and Lillian Lee. 2005.
\newblock Seeing stars: Exploiting class relationships for sentiment
  categorization with respect to rating scales.
\newblock In {\em Proceedings of ACL\/}. pages 115--124.

\bibitem[{Pennington et~al.(2014)Pennington, Socher, and
  Manning}]{Pennington:2014emnlp}
Jeffrey Pennington, Richard Socher, and Christopher~D. Manning. 2014.
\newblock Glo{V}e: Global vectors for word representation.
\newblock In {\em Proceedings of EMNLP\/}.

\bibitem[{Perfetti(1998)}]{Perfetti:1998dp}
Charles~A Perfetti. 1998.
\newblock The limits of co-occurrence: Tools and theories in language research.
\newblock {\em Discourse Processes\/} 25(2\&3):363---377.

\bibitem[{Pham et~al.(2015)Pham, Kruszewski, Lazaridou, and
  Baroni}]{Pham:2015acl}
Nghia~The Pham, German Kruszewski, Angeliki Lazaridou, and Marco Baroni. 2015.
\newblock Jointly optimizing word representations for lexical and sentential
  tasks with the c-phrase model.
\newblock In {\em Proceedings of ACL\/}.

\bibitem[{Reed et~al.(2016)Reed, Akata, Lee, and Schiele}]{Reed:2016cvpr}
S.~Reed, Z.~Akata, H.~Lee, and B.~Schiele. 2016.
\newblock Learning deep representations of fine-grained visual descriptions.
\newblock In {\em Proceedings of CVPR\/}.

\bibitem[{Riordan and Jones(2011)}]{Riordan:2011tcs}
Brian Riordan and Michael~N Jones. 2011.
\newblock Redundancy in perceptual and linguistic experience: Comparing
  feature-based and distributional models of semantic representation.
\newblock {\em Topics in Cognitive Science\/} 3(2):303--345.

\bibitem[{Saxe et~al.(2014)Saxe, McClelland, and Ganguli}]{Saxe:2014iclr}
Andrew~M. Saxe, James~L. McClelland, and Surya Ganguli. 2014.
\newblock Exact solutions to the nonlinear dynamics of learning in deep linear
  neural networks.
\newblock In {\em International Conference on Learning Representations
  (ICLR)\/}.

\bibitem[{Socher et~al.(2013)Socher, Perelygin, Wu, Chuang, Manning, Ng, Potts
  et~al.}]{Socher:2013emnlp}
Richard Socher, Alex Perelygin, Jean~Y Wu, Jason Chuang, Christopher~D Manning,
  Andrew~Y Ng, Christopher Potts, et~al. 2013.
\newblock Recursive deep models for semantic compositionality over a sentiment
  treebank.
\newblock In {\em Proceedings of EMNLP\/}. pages 1631---1642.

\bibitem[{Sutskever et~al.(2014)Sutskever, Vinyals, and
  Le}]{Sutskever:2014nips}
I.~Sutskever, O.~Vinyals, and QV. Le. 2014.
\newblock Sequence to sequence learning with neural networks.
\newblock In {\em Proceedings of NIPS\/}.

\bibitem[{Tarski(1944)}]{Tarski:1944ppr}
Alfred Tarski. 1944.
\newblock The semantic conception of truth: and the foundations of semantics.
\newblock {\em Philosophy and phenomenological research\/} 4(3):341--376.

\bibitem[{Turney and Pantel(2010)}]{Turney:2010jair}
Peter~D. Turney and Patrick Pantel. 2010.
\newblock From {F}requency to {M}eaning: vector space models of semantics.
\newblock {\em Journal of Artifical Intelligence Research\/} 37(1):141--188.

\bibitem[{van~den Oord et~al.(2016)van~den Oord, Kalchbrenner, Espeholt,
  kavukcuoglu, Vinyals, and Graves}]{Oord:2016nips}
Aaron van~den Oord, Nal Kalchbrenner, Lasse Espeholt, koray kavukcuoglu, Oriol
  Vinyals, and Alex Graves. 2016.
\newblock Conditional image generation with pixelcnn decoders.
\newblock In {\em Proceedings of NIPS\/}.

\bibitem[{Vendrov et~al.(2015)Vendrov, Kiros, Fidler, and
  Urtasun}]{Vendrov:2015iclr}
Ivan Vendrov, Ryan Kiros, Sanja Fidler, and Raquel Urtasun. 2015.
\newblock Order-embeddings of images and language.
\newblock In {\em International Conference on Learning Representations
  (ICLR)\/}.

\bibitem[{Vinyals et~al.(2015)Vinyals, Toshev, Bengio, and
  Erhan}]{Vinyals:2015cvpr}
O.~Vinyals, A.~Toshev, S.~Bengio, and D.~Erhan. 2015.
\newblock Show and tell: {A} neural image caption generator.
\newblock In {\em Proceedings of CVPR\/}.

\bibitem[{Wiebe et~al.(2005)Wiebe, Wilson, and Cardie}]{Wiebe:2005lrec}
Janyce Wiebe, Theresa Wilson, and Claire Cardie. 2005.
\newblock Annotating expressions of opinions and emotions in language.
\newblock {\em Language resources and evaluation\/} 39(2):165--210.

\bibitem[{Wieting et~al.(2016)Wieting, Bansal, Gimpel, and
  Livescu}]{Wieting:2016iclr}
John Wieting, Mohit Bansal, Kevin Gimpel, and Karen Livescu. 2016.
\newblock Towards universal paraphrastic sentence embeddings.
\newblock In {\em International Conference on Learning Representations
  (ICLR)\/}.

\end{thebibliography}
\bibliographystyle{acl_natbib}


\end{document}